\documentclass[twocolumn]{svjour3}          
\smartqed  
\usepackage{graphicx}
\usepackage[hidelinks]{hyperref}
\usepackage{amsmath,amssymb,mathtools}

\usepackage{times}    

\usepackage[labelfont=bf]{caption}
\usepackage{subcaption}

\newtheorem{assumption}{Assumption}

\newcommand{\real}{\mathbb{R}}

\newcommand{\ex}{\mathbb{E}}
\newcommand{\Imspace}{\mathcal{I}(\mathcal{D})}
\newcommand{\VF}{\mathfrak{X}(\mathcal{D})}
\newcommand{\dVF}{\mathfrak{X}^*(\mathcal{D})}
\newcommand{\cF}{\mathcal{F}}
\newcommand{\sepda}{(\ref{eq:StochEPDiff},~\ref{eq:StochAdvect})}

\newcommand{\ad}{\operatorname{ad}}

\newcommand{\mermaid}{\href{https://mermaid.readthedocs.io/en/latest/index.html}{\textsf{mermaid}} }
\newcommand{\pytorch}{\href{https://pytorch.org/}{\textsf{PyTorch}} }
\newcommand{\torchdiffeq}{\href{https://github.com/rtqichen/torchdiffeq}{\textsf{torchdiffeq}} }

\DeclareMathOperator{\Div}{div}
\DeclareMathOperator{\Diff}{Diff}
\DeclareMathOperator*{\argmin}{arg\,min}
\DeclareMathOperator{\Id}{Id}
\DeclarePairedDelimiterX{\inp}[2]{\langle}{\rangle}{#1, #2}
\DeclarePairedDelimiterX{\mom}[1]{\langle}{\rangle}{#1}

\journalname{Journal of Mathematical Imaging and Vision}

\begin{document}

\title{Moment evolution equations and moment matching for stochastic image EPDiff}

\titlerunning{Stochastic Image Deformation}        

\author{
Alexander Mangulad Christgau \and Alexis Arnaudon \and Stefan~Sommer 
}

\authorrunning{A.M. Christgau \and A. Arnaudon \and S. Sommer} 

\institute{Alexander Mangulad Christgau \at
				Department of Mathematical Sciences, University of Copenhagen, Denmark \\       				\email{amc@math.ku.dk}
			\and
		Alexis Arnaudon \at
    		Department of Mathematics, Imperial College, London, UK.
		    \email{alexis.arnaudon@imperial.ac.uk}
			\and
        Stefan Sommer \at
			Department of Computer Science, University of Copenhagen, Denmark \\ \email{sommer@di.ku.dk}
}

\date{Received: date / Accepted: date}

\maketitle

\begin{abstract}
Models of stochastic image deformation allow study of time-continuous stochastic effects transforming images by deforming the image domain. Applications include longitudinal medical image analysis with both population trends and random subject specific variation.
Focusing on a stochastic extension of the LDDMM models with evolutions governed by a stochastic EPDiff equation, we use moment approximations of the corresponding Itô diffusion to construct estimators for statistical inference in the full stochastic model. We show that this approach, when efficiently implemented with automatic differentiation tools, can successfully estimate parameters encoding the spatial correlation of the noise fields on the image.

\end{abstract}

\section{Introduction} \label{intro}
The Large deformation diffeomorphic metric mapping (LDDMM) framework was developed with the intention to model deformations of images and shapes, driven by applications in fields including medical imaging and biology. LDDMM models shape and image evolution as a gradual process induced by time-continuous paths $\phi_t$ of deformations. The model is therefore naturally applicable to longitudinal studies to investigate shape evolution of human organs during child development, natural aging or disease processes.

\begin{figure*}[htb!]
  \centering
  \includegraphics[width = .75\linewidth]{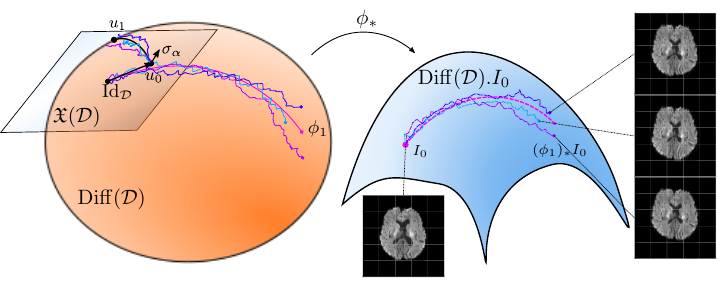}
  \caption{(left) Illustrations of some random sample paths on the diffeomorphism group $\Diff(\mathcal{D})$ induced by a random motion of vector fields governed by the stochastic EPDiff equation \eqref{eq:StochEPDiff}. (center)
  Starting with an image $I_0$, e.g. a brain MR-slice, each path of diffeomorphisms deforms the image by the pushforward-action. (right) Some resulting samples of the endpoint image $(\phi_1)_*I_0 = I_0\circ \phi_1^{-1}$ are shown on the right.}
  \label{fig:StochasticOrbit}
\end{figure*}

To model subject-specific deviations from a population-mean deformation, it is natural to incorporate noise into the LDDMM framework. One approach is to consider random variation in the initial velocity $u_0 = \partial_t \phi_t|_{t=0}$ as in e.g. the random orbit model \cite{RandomOrbit,tang2013bayesian} or Bayesian principal geodesic analysis \cite{zhang2015bayesian}. 
Instead of assuming that the entire variation is the result of a random event at the beginning, \cite{trouve_shape_2012,vialard_extension_2013,marsland_langevin_2017,arnaudon2019geometric} proposed to include time-continuous noise. A general framework for stochastic shape analysis with the variability incorporated as random variation over the entire evolution is proposed in \cite{arnaudon2019geometric} by using \emph{noise fields} $\sigma_1,\ldots, \sigma_p$, to perturb the deformation $\phi_t$, thus creating the stochastic flow. The framework and examples of its effect on medical images is illustrated in Figure~\ref{fig:StochasticOrbit}. Estimation of noise in this framework was considered in \cite{arnaudon2019geometric} for landmark data. Noise estimation for images is inherently more difficult due to the infinite dimensional nature of the observations and the non-linear coupling between deformation and image. 

In this paper, we develop a new approach for estimation of noise in the stochastic framework from image data. Specifically, we derive moment equations of the stochastic evolution in image and momentum space, and we implement these in a modern image registration framework allowing automatic differentiation of the moment equations and subsequent optimization. Using the derived estimators, we perform simulation studies demonstrating the ability of the scheme to estimate unknown parameters of the stochastic noise. 

The paper builds on initial steps for noise estimation from image deformation models using strings \cite{arnaudon2018string} and the Fourier space moment approximations in the preprint \cite{kuhnel2018stochastic}.
The paper thus presents the following contributions:
\begin{enumerate}
    \item We derive moment equations for the stochastic EPDiff and advection equations of images (SEPDA), and propose a first-order approximation.
    \item We construct estimators based on matching of first-order moments of the images.
    \item We show how the scheme can be implemented in a modern image registration framework allowing automatic differentiation through the approximated moment equations.
    \item We demonstrate the ability of the methods to estimate noise fields with simulated data, both in a correctly specified model and with misspecification.
\end{enumerate}

\subsection{Plan of Paper}
In Section~\ref{sec:Background}, we briefly describe the LDDMM framework and its stochastic generalization from \cite{arnaudon2019geometric} in the special case of image deformation. In Section~\ref{sec:estimation}, we first derive the moment equations for the momentum and image flow $(m_t,I_t)$. We then use approximations of these to define a new procedure for parameter estimation in the stochastic framework based on the method of moments. In Section~\ref{sec:implementation}, we discuss the numerical implementation which is used investigate the moment estimators with simulation studies. In Section~\ref{sec:results}, the results of the simulation studies are presented and discussed. We end the paper with concluding remarks.

\section{Background}
\label{sec:Background}
We introduce the LDDMM framework in the context of image registration. A detailed treatment can be found in \cite{younes2010shapes}, and, with a perspective from geometric mechanics, in \cite{bruveris2011momentum,bruveris2015geometry}.

Our main objects of interest are grayscale images, which are given as maps $\mathcal{D} \to \real$ defined on an image domain $\mathcal{D} \subseteq \real^d$. We denote the space of images by $\Imspace$. Our prototypical example of an image domain is $\mathcal{D}=(0,1)^2$, although the theory naturally extends to general domains including 3D-images.
Given a \emph{source image} $I_0$ and a \emph{target image} $T$, the informal objective of LDDMM image registration is: Find a ``deformation'' that transforms $I_0$ into an approximation of $T$, subject to minimal ``complexity''. We proceed to give a more rigorous explanation of this objective.

A \textit{deformation} of images is modeled by a path of diffeomorphisms $\phi = (\phi_t)_{t\in[0,1]} \subset \Diff (\mathcal{D})$ starting at the identity map $\phi_0 = \Id_{\mathcal{D}}$, where $\Diff (\mathcal{D})$ denotes the group of diffeomorphisms on $\mathcal{D}$. A deformation is further assumed to be generated as the flow of a time-dependent vector field, 
$$
    u = (u_t)_{t\in[0,1]} \subset \VF\coloneqq C^\infty(\mathcal{D},\real^d),
$$
via the \emph{reconstruction equation}
\begin{equation}\label{eq:reconstruct}
    \partial_t \phi_t = u_t \circ \phi_t, \qquad t\in [0,1].
\end{equation}
This equation and every subsequent equation of mappings can be interpreted by pointwise evaluation for each point in $x\in \mathcal{D}$. The name of \eqref{eq:reconstruct} is due to the one-to-one correspondence between $u$ and $\phi$, which is ensured by the fundamental theorem of time-dependent flows \cite[Thm. 9.48]{lee2013smooth}. We therefore also refer to $u$ as a \emph{velocity field}. For notational ease, we will omit the subscript $t\in[0,1]$ and simply write $u=(u_t)$, and similarly, $\phi = (\phi_t)$.

Given a deformation $(\phi_t)$, we consider the path of images starting at $I_0$ and evolving by the pushforward-action:
$I_t \coloneqq (\phi_t)_{*}I_0 = I_0 \circ \phi_t^{-1}$.
To measure how well $(\phi_t)$ transforms $I_0$ into $T$, we define the \emph{total energy} functional $E$ by
\begin{equation}\label{eq:energyfunctional}
    E(u) = \frac{1}{2} \int_0^1\|u_t\|_{\VF}^2dt + \frac{1}{2\lambda^2}\|I_0 \circ \phi_1^{-1}-T\|_{L^2(\mathcal{D})}^2.
\end{equation}
Here $\lambda>0$ is a given trade-off parameter and the norm on $\VF$ is, by assumption, induced by an inner product $\inp{u}{v}_{\VF} \coloneqq  \inp{u}{Lv}_{L^2(\mathcal{D},\real^3)}$ for some positive self-adjoint operator $L\colon \VF \to \VF$. Note that we have used the one-to-one correspondence to write $E$ as a functional of $u=(u_t)$ instead of $\phi = (\phi_t)$.

The first term of $E$ is the \emph{kinetic energy}, which measures the regularity of the deformation~$\phi$. 
The second term measures the dissimilarity between the endpoint image $I_1=I_0\circ \phi_1^{-1}$ and the target image $T$. 
The LDDMM image registration objective is to minimize the energy functional $E$ over all deformations. Thus the objective seeks a deformation compromising between kinetic energy and dissimilarity.

\begin{figure}[htb!]
  \centering
  \includegraphics[width=\linewidth]{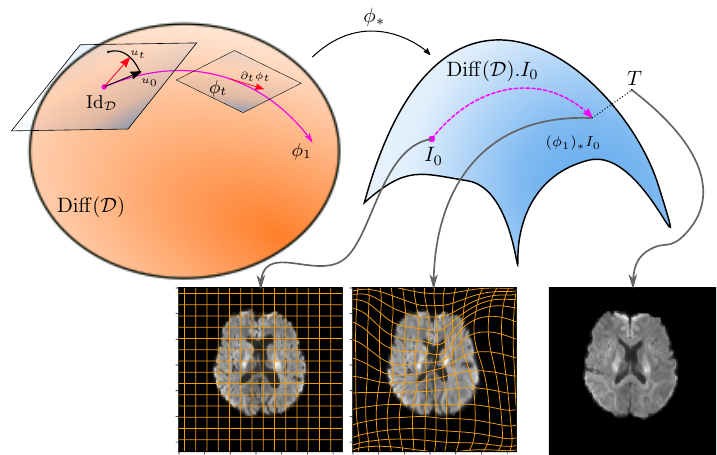}
  \caption{An illustration of the deformation $t \mapsto \phi_t\in \Diff(\mathcal{D})$ acting on the brain MR-slice $I_0$. The deformation transforms $I_0$ into an image $I_1 = (\phi_1)_* I_0$ being close to $T$, while also maintaining a low kinetic energy determined from its speed $\|u_t\|$. Observe for example that the lateral ventricles are contracted from $I_0$ to $I_1$ to be more reminiscent of those in $T$.}
  \label{fig:image_registration}
\end{figure}

Below we discuss the dynamics of a minimizer of LDDMM objective. Based on these dynamics, it is possible to match images as shown in Figure~\ref{fig:image_registration}, which illustrates image registration with two brain MR-slices. 

\subsection{The EPDiff Equation}
In the seminal paper \cite{beg2005computing}, Faisal Beg implemented the first algorithm for minimizing the LDDMM objective. The algorithm computes a functional derivative of $E$ to iteratively update the deformation as a form of steepest descent. This has inspired several approaches of using the calculus of variations in one form or another to produce other governing equations and algorithms for LDDMM \cite{bruveris2015geometry}. We focus on a particular form of the dynamics, stating that $u$ is a stationary point of the energy $E$ defined in equation \eqref{eq:energyfunctional} if and only if it satisfies the \emph{EPDiff equation}:
\begin{align}\label{eq:EPDiff}
    \partial_t m_t + Dm_t.u_t + (Du_t)^T.m_t 
    + \Div(u_t)m_t &= 0. \\
    m_t &= Lu_t, \nonumber
\end{align}
Here $Dv(x)=(\frac{\partial v^i}{\partial x^j})_{j=1,\ldots,d}^{i=1,\ldots,d}$ is understood as the Jacobian with respect to the spatial coordinates for any $v\in \VF$ and $x\in \mathcal{D}$. Derivations of the EPDiff equation can be found in \cite{bruveris2015geometry,holm2009geometric,younes2010shapes}.

An immediate application of the EPDiff equation is image registration by the \textit{shooting method}. Given an initial velocity $u_0$, the EPDiff equation can be solved to recover a stationary path $u = (u_t)_{t\in[0,1]}$ from which $E$ can be computed. 
The practical details and various implementation strategies of this method are discussed for example in~\cite{allassonniere2005geodesic,trouve_shape_2012,niethammer2019metric}.

Instead of first computing the deformation $(\phi_t)$ from the reconstruction equation, the path of images $(I_t)$ may also be computed directly from $(u_t)$. 
Indeed, by differentiating $I_t = I_0 \circ \phi_t^{-1}$ and using the reconstruction equation~\eqref{eq:reconstruct}, one arrives at the \emph{image advection equation}:
\begin{equation}\label{eq:adveceq}
    \partial_t I_t = - u_t\cdot \nabla I_t\, ,
\end{equation}
where $\nabla$ denotes the gradient with respect to the spatial coordinates \cite{bruveris2015geometry}.

In practice, the operator $L\colon \VF \to \dVF$ is specified from the Green's Kernel $K\colon \mathcal{D}\times \mathcal{D} \to \real^d$, given by $LK(\cdot,y) = \delta(\cdot - y)$, where $\delta$ is the Dirac distribution. 
For spatially invariant kernels of the form $K(x,y)=k(x-y)\mathbf{1}_d$, for some $k \colon \mathcal{D}^2 \to \real$, $K$ is related to $L$ by $L^{-1}m = k*m$, where $*$ denotes convolution. For the simulation studies introduced later, a multi-Gaussian kernel of the form $k(x) =\sum_{i=1}^n w_i \exp(- x^2/\sigma_i^2)$ was used following~\cite{niethammer2019metric}.

Before introducing stochasticity, we remark that the EPDiff equation has a deeper geometric interpretation. 
Indeed, the velocity field $u$, being a minimizer of $E$, is a geodesic with respect to $\inp{\cdot}{\cdot}_{\VF}$, where $\inp{\cdot}{\cdot}_{\VF}$ can be interpreted as a Riemannian metric on $\Diff(\mathcal{D})$ \cite{bruveris2015geometry}. 
In fact, this observation is part of a more general discovery by Vladimir Arnold~\cite{arnold1966geometrie}, who connected the dynamics of certain mechanical systems with geodesics on Lie groups. 
The EPDiff equation can therefore be seen as a special case of the \emph{Euler-Poincaré (Euler-Arnold) equation}:
\begin{equation}
    \partial_t m_t + \ad_{u_t}^*m_t = 0, \qquad m_t = Lu_t\, ,
\end{equation}
where $\ad^*$ is the so-called the \emph{coadjoint operator} from Lie theory, which we will use as a shorthand notation for
\begin{equation} \label{eq:coadjoint}
    \ad_{u}^*m = Dm.u+ (Du)^T.m + \Div(u)m.
\end{equation}
Likewise, we refer to $m_t = Lu_t$ as the \emph{momentum} associated to $u_t$. 
This general mechanical perspective is the starting point for the stochastic generalization introduced in~\cite{arnaudon2019geometric}, but we omit the precise details since our focus is on image deformations.

\subsection{Perturbing the Reconstruction Equation}

To account for random variations in deformations, we now describe the stochastic framework proposed in~\cite{arnaudon2019geometric}. Relevant material on stochastic calculus and stochastic differential equations (SDE's) can be found in \cite{oeksendal2010stochastic,karatzas2014brownian,evans2012introduction,van1992stochastic,kunita2019stochastic}.

The framework is derived from perturbing the flow of the time-dependent velocity field $(u_t)$ by stochastic noise. To this end, let $\sigma_1,\ldots,\sigma_p \in \VF$ be a collection of vector fields which we refer to as \textit{noise fields}.
The noise will be modeled with a $p$-dimensional Wiener processes $W = (W_t^\alpha)_{t\in[0,1]}^{\alpha=1,\ldots,p}$ defined on a universal background probability space $(\Omega, \cF, P)$ and with independent components $W^\alpha = (W_t^\alpha)_{t\geq 0}$. 
With this noise, we perturb the reconstruction equation \eqref{eq:reconstruct} to obtain the \emph{Stratonovich SDE}
\begin{equation}\label{eq:stochReconstruction}
    d\phi_t(x) = u_t(\phi_t(x) )dt + \sum_{\alpha =1}^p \sigma_\alpha(\phi_t(x)) \circ dW_t^\alpha\, ,
\end{equation}
parameterized over $x \in \mathcal{D}$. The SDE is a shorthand for its corresponding stochastic integral equation,
\begin{equation*}
    \phi_t(x) = \int_0^t u_s(\phi_s(x))ds + \sum_{\alpha =1}^p 
    \int_0^t \sigma_\alpha(\phi_s(x)) \circ dW_s^\alpha\, ,
\end{equation*}
where the right-most integrals are Stratonovich integrals \cite{oeksendal2010stochastic,karatzas2014brownian,stratonovich1966new,kunita2019stochastic}. In particular, $\circ dW_t^\alpha$ should not be confused with functional composition. Intuitively, the SDE asserts that the deformation is guided by the velocity field plus a random amount of attraction in the direction of each noise field.
As such, the deformation $(\phi_t)$ is a time-continuous stochastic process on $\Diff(\mathcal{D})$. 

In principle, we could also perturb the dynamics with an \emph{Itô integral}. However, when modeling mechanical systems with external noise, it is generally more suitable to incorporate stochasticity with the Stratonovich formulation, see e.g. the ``Itô vs Stratonovich dilemma'' in \cite{van1992stochastic}. One reason is that the Stratonovich formulation obeys the classical chain rule rather than \emph{Itô's lemma}. The choice of the Stratonovich formulation is also motivated by \cite{holm2015variational}, which formulates stochastic fluid dynamics in terms of Stratonovich perturbations.

The stochastic perturbations in \eqref{eq:stochReconstruction} are given in Eulerian coordinates. This is in line with the LDDMM framework which, because of the right-invariance of the metric, is also Eulerian. The metric and stochastic structure are compatible allowing reduction to the specific shape space comparable to LDDMM and thus leading to stochastic versions of the Euler-Poincar\'e equations. LDDMM can be rephrased as left-invariant allowing the reference frame and metric to follow deformed structures~\cite{schmahLeftInvariantMetricsDiffeomorphic2013}. Applying a similar change of reference frame may be possible for the noise. Another possibility for focusing the noise on specific structures is to use a construction similar to  region specific LDDMM \cite{RDDMM}.

We can define the perturbed energy functional $\tilde E$ by the same expression as in \eqref{eq:energyfunctional}, but where $\phi_1$ is obtained from the perturbed reconstruction equation \eqref{eq:stochReconstruction} instead:
\begin{align*}
    \tilde E(u) = \frac{1}{2} \int_0^1\|u_t\|_{\VF}^2dt + \frac{1}{2\lambda^2}\|I_0 \circ \phi_1^{-1}-T\|_{L^2(\mathcal{D})}^2,\\
    \phi_1 \text{ is a solution to } \eqref{eq:stochReconstruction}\, .
\end{align*}
Proposition 2.4 of \cite{arnaudon2019geometric} asserts that if $u_t$ is a critical point of the perturbed energy $\tilde E$, then it satisfies the \emph{stochastic Euler-Poincaré equation}:
\begin{equation}\label{eq:StochEulerPoincare}
    dm_t + \ad_{u_t}^* m_t dt 
    + \sum_{\alpha=1}^p \ad_{\sigma_\alpha}^*u_t \circ W_t^\alpha = 0\, .
\end{equation}
In~\cite{arnaudon2019geometric}, the above dynamical equation was derived for general shapes rather than images. In our case, the coadjoint operator $\ad^*_{\sigma_\alpha}$ is given by \eqref{eq:coadjoint} and can be inserted in the stochastic Euler-Poincaré equation \eqref{eq:StochEulerPoincare} to obtain the \emph{stochastic EPDiff equation} \cite{arnaudon2018string}:
\begin{align} \label{eq:StochEPDiff}
    dm_t = 
    - &\left(Dm_t.u_t + (Du_t)^T.m_t + \Div(u_t)m_t\right)dt \nonumber \\
    -&\left(Dm_t .\sigma_\alpha 
    +(D\sigma_\alpha)^T.m_t +\Div(\sigma_\alpha)m_t\right)\circ dW_t^\alpha\, ,
\end{align}
where we have used the Einstein summation convention on the index $\alpha = 1,\ldots, p$. 

Since the Stratonovich noise is added linearly to the reconstruction equation, and since the advection equation is linear in $u_t$, the resulting dynamics of the image evolution is becomes
\begin{equation}\label{eq:StochAdvect}
    dI_t = -\nabla I_t \cdot u_t dt - \left(\nabla I_t \cdot \sigma_\alpha\right) \circ dW_t^\alpha.
\end{equation}
We refer to \eqref{eq:StochAdvect} as the \emph{stochastic advection equation}, see Equation 3.4 in \cite{arnaudon2018string} for a derivation. In the following we refer to the stochastic EPDiff \eqref{eq:StochEPDiff} and the stochastic advection equation \eqref{eq:StochAdvect} collectively as the \textit{SEPDA} equations. We avoid technical questions regarding the existence and regularity of solutions but only assume the following.
\begin{assumption}\label{asm:solution}
    There exists an $\VF \times \Imspace$-valued process $(u_t,I_t)_{t\in[0,1]}$, with $(u_t)\in L^2([0,1]\times\Omega, \VF)$ and $(I_t)\in L^2([0,1]\times\Omega, \Imspace)$, which is a strong solution the SEPDA equations. That is, almost surely
    \begin{align*}
        I_t = I_0 
            - \int_0^t \nabla I_s \cdot u_s ds
            - \sum_{\alpha=1}^p \int_0^t \left(\nabla I_s \cdot \sigma_\alpha\right) \circ dW_s^\alpha\, , 
    \end{align*}
    for all $t\in [0,1]$, and similarly, $(u_t)$ obeys the stochastic EPDiff equation \eqref{eq:StochEPDiff}.
\end{assumption}

If desired, we may consider the SEPDA equations \sepda{} as a starting point of the stochastic shape framework of~\cite{arnaudon2019geometric} specialized to images.
Figure~\ref{fig:StochasticOrbit} illustrates different sample realizations of the dynamics defined by the SEPDA equations, with the initial image $I_0$ being a brain MR-slice.

\section{Parameter Estimation}
\label{sec:estimation}
A priori, the SEPDA equations define a non-parametric model for stochastic image deformation in the sense that there is an infinite-dimensional freedom in the choice of noise fields. A particular parametric submodel can be specified by parametrizing the noise fields $\sigma_1,\ldots,\sigma_p$, i.e., specifying a map of the form:
\begin{align}\label{eq:parametricsubmodel}
    \Sigma \colon \Theta& \longrightarrow \VF^p,  \nonumber \\
    \qquad \theta& \longmapsto \Sigma(\theta) = (\sigma_1(\theta), \ldots, \sigma_p(\theta)),
\end{align}
where $\theta\in \Theta \subseteq \real^q$ is the \emph{noise field parameter}. Note that $p$ is considered as a hyperparameter for simplicity. To estimate the noise field parameters we proceed with the method of moments.

\subsection{Moment Equations}
\label{sub:mom}
Let $(m_t,I_t)_{0\leq t\leq 1}$ be solutions to the SEPDA equations \sepda{}. The moment of a stochastic process of maps on $\mathcal{D}$ is defined pointwise and denoted by $\langle\cdot \rangle$, e.g.,
\begin{align}
    \mom{I_t} \colon \mathcal{D} &\longrightarrow \real, \\
 \mom{I_t}(x) &= \ex I_t(x) = \int_{\Omega} I_t(\omega)(x)dP(\omega).
\end{align}

Note that integrability is ensured by 
Assumption~\ref{asm:solution}.
To compute the moments, we reformulate the SEPDA equations as Itô SDE's. Since Itô integrals are martingales, the noise terms of Itô the SDE's will vanish in expectation. We briefly summarize the Itô-Stratonovich conversion (see for instance ~\cite{oeksendal2010stochastic} for more details).

Let $W = (W_t^\alpha)_{t\in [0,1]}^{\alpha = 1,\ldots,p}$ be a $p$-dimensional Wiener process, let $a\colon \real^n \to \real^n$ and let $b=(b_\alpha)\colon \real^n \to \real^{n\times p}$ with $b_\alpha \colon \real^n \to \real^n$. Then with sufficient regularity conditions on $a$ and $b$, an $n$-dimensional stochastic process $X = (X_t)$ is a solution to the Stratonovich SDE 
\begin{equation*}
    dX_t = a(X_t) dt + b(X_t) \circ dW_t
\end{equation*}
if and only if it is a solution the Itô SDE
\begin{equation*}
    dX_t = \big[a(X_t) + \frac{1}{2}c(X_t)\big]dt + b(X_t)\cdot dW_t,
\end{equation*}
where the term $\frac{1}{2}c(X_t)$ is the \textit{Itô-Stratonovich correction term} given by
\begin{align}
    c(x) &= \sum_{\alpha=1}^p Db_\alpha(x).b_\alpha(x). \label{eq:correctionterm}
\end{align}
Consider the special case where each $b_\alpha(x) = B_\alpha .x$ is a linear map with $B_\alpha \in \real^{n\times n}$. Then the Jacobian $Db_\alpha = B_\alpha$ corresponds to the linear map itself, and hence the Itô-Stratonovich correction term is $c(x) = \sum_{\alpha=1}^p Db_\alpha(x).b_\alpha(x) = \sum_{\alpha=1}^p b_\alpha(b_\alpha(x))$.

When the SDE is replaced by an SPDE, the Itô-Stratonovich correction term is formally the same as \eqref{eq:correctionterm}, but the derivative $D$ should be interpreted as a Fréchet derivative since $b$ is an operator on a normed function space in this case. See Section 4.5.2 in \cite{duan2014effective} for a derivation. Since the coadjoint operator $\ad_{\sigma_\alpha}^*\colon \dVF \to \dVF$ and the \emph{fundamental vector fields} \cite{bruveris2015geometry}
\begin{equation}
 \zeta_{\sigma_\alpha} \colon \Imspace \longrightarrow \Imspace, \qquad I\longmapsto\nabla I \cdot \sigma_\alpha,   
\end{equation}
are linear maps of normed spaces, their Fréchet derivatives are equal to the maps themselves. Hence the Itô-Stratonovich correction terms of the SEPDA equations \sepda{} are
\begin{align}
    c_{\operatorname{EPDiff}}(m_t) &=  \sum_{\alpha=1}^p
    \ad_{ \sigma_\alpha }^* (\ad_{\sigma_\alpha}^*  m_t) \label{eq:EPDiffCorrect},\\
    c_{\operatorname{Advect}}(I_t) &= \sum_{\alpha=1}^p
    \zeta_{ \sigma_\alpha } (\zeta_{\sigma_\alpha}  (I_t)) = \sum_{\alpha=1}^p
    \nabla(\nabla I_t\cdot \sigma_\alpha) \cdot \sigma_\alpha. \nonumber
\end{align}
In principle, we could write \eqref{eq:EPDiffCorrect} explicitly using the expression for the coadjoint operator in \eqref{eq:coadjoint}. However, for brevity we choose to work with the more compact $\ad^*$ henceforth.

The corresponding Itô formulations of the SEPDA equations are therefore
\begin{align}
    dm_t  &=
    \Big[-\ad_{ u_t}^*  m_t 
    +\frac{1}{2} \sum_{\alpha=1}^p
    \ad_{ \sigma_\alpha }^* (\ad_{\sigma_\alpha}^*  m_t)\Big]dt + \text{Itô noise}, \nonumber \\ 
    dI_t 
    &= \Big[-\nabla  I_t \cdot u_t
    +
    \frac{1}{2} \sum_{\alpha=1}^p
    \nabla(\nabla I_t\cdot \sigma_\alpha) \cdot \sigma_\alpha
    \Big]dt \nonumber \\ 
    & \qquad + \text{Itô noise.}
    \label{eq:ItoFormulations}
\end{align}
Taking the expectation of the momentum equation and using Fubinis theorem yields:
\begin{align*}
    \mom{m_t}
    &=
        \ex \int_0^t \Big[-\ad_{u_s}^* m_s
        +\frac{1}{2} \sum_{\alpha=1}^p
        \ad_{ \sigma_\alpha }^* (\ad_{\sigma_\alpha}^*  m_s)\Big]ds + 0 \\
    &=
        \int_0^t \ex \Big[-\ad_{u_s}^* m_s
        +\frac{1}{2} \sum_{\alpha=1}^p
        \ad_{ \sigma_\alpha }^* (\ad_{\sigma_\alpha}^*  m_s)\Big]ds \\
    &=
        \int_0^t \Big[-\mom{\ad_{u_s}^* m_s}
        +\frac{1}{2} \sum_{\alpha=1}^p
        \mom{\ad_{ \sigma_\alpha }^* (\ad_{\sigma_\alpha}^*  m_s)}\Big]ds \\
    &=
        \int_0^t \Big[-\mom{\ad_{u_s}^* m_s}
        +\frac{1}{2} \sum_{\alpha=1}^p
        \ad_{ \sigma_\alpha }^* (\ad_{\sigma_\alpha}^* \mom{m_s})\Big]ds.
\end{align*}
In the last equality, we have interchanged the expectation with the linear differential operators $\ad_{\sigma_\alpha}^*$, $\alpha=1,\ldots,p$, which is justified by Leibniz integral rule. Thus we conclude that
\begin{align} \label{eq:momentummoment}
    \partial_t\mom{m_t}  &=
    -\mom{\ad_{ u_t}^*  m_t} 
    +\frac{1}{2} \sum_{\alpha=1}^p
    \ad_{ \sigma_\alpha }^* 
    (\ad_{\sigma_\alpha}^* \mom{m_t}).
\end{align}
With an analogous argument, taking the expectation of the stochastic advection equation yields
\begin{align} \label{eq:imagemoment}
    \partial_t \langle I_t \rangle
    &=- \langle \nabla I_t\cdot u_t \rangle
    +\frac{1}{2} \sum_{\alpha=1}^p
    \nabla(\nabla \mom{I_t}\cdot \sigma_\alpha) \cdot \sigma_\alpha.
\end{align}
Unfortunately, equations \eqref{eq:momentummoment} and \eqref{eq:imagemoment} cannot be forward integrated due to the unknown terms $\mom{\ad_{ u_t}^*  m_t}$ and $\langle \nabla I_t\cdot u_t \rangle$ appearing on the right-hand side.
We therefore resort to the coarse approximations
\begin{align} 
    \partial_t\mom{m_t} 
    &\approx -\ad_{\mom{ u_t}}^* \mom{m_t} 
    +\frac{1}{2} \sum_{\alpha=1}^p
    \ad_{ \sigma_\alpha }^*
    (\ad_{\sigma_\alpha}^*  \mom{m_t}), \label{eq:momentequation1} \\
    \partial_t \langle I_t \rangle
    &\approx - \nabla \langle I_t \rangle \cdot \langle u_t\rangle
    +\frac{1}{2} \sum_{\alpha=1}^p 
    \nabla(\nabla \mom{I_t}\cdot \sigma_\alpha) \cdot \sigma_\alpha. \label{eq:momentequation2}
\end{align}
Below we explain the rationale behind the approximations, but note first that the system can be forward integrated as it can be expressed purely in terms of $\mom{m_t}$, $\mom{I_t}$ and the noise fields. This follows from the relation $\mom{u_t} = k * \mom{m_t}$, which can be shown using Fubinis theorem: 
\begin{align*}
    \mom{u_t}(x)
    &= \ex \int_\mathcal{D} k(x-y)m_t(y) dy \\
    &= \int_\mathcal{D} k(x-y)\ex m_t(y) dy 
    = (k * \mom{m_t})(x).
\end{align*}
The approximations \eqref{eq:momentequation1} and \eqref{eq:momentequation2} are based on replacing the moments
\begin{equation*}
    \mom{Dm_t. u_t}, \quad
    \mom{(Du_t)^T. m_t}, \quad
    \mom{\Div(u_t) m_t}, \quad
    \mom{\nabla I_t \cdot u_t}
\end{equation*}
with the products of each moment, for example,
$$
    \mom{Dm_t. u_t} \approx \mom{Dm_t}.\mom{u_t}.
$$
While there is a clear description of the relationship between $u_t$ and $m_t$, it is not a priori clear how to describe their correlation structure. The smaller the perturbation is, i.e., the smaller the magnitude of the noise fields is, the smaller we expect the correlation to be. In the limit, when the noise fields are the zero vector field, the approximations are exact and we note that \eqref{eq:momentequation1} and \eqref{eq:momentequation2} simplify to the regular EPDiff and advection equations.  

The approximations are a first order example of the \emph{cluster expansion method} \cite{kira2011semiconductor}. In \cite{arnaudon2019geometric}, a second order approximation was computed for the moment equations of stochastic landmark dynamics. A second order approximation might also be possible for images, although the computations would be considerably more technical due to the infinite-dimensional nature of this setting. The inclusion of higher order moments will thus be left as a topic for future research. 

\subsection{Constructing an Estimator}
Consider now a parametric model of the noise fields $(\sigma_1(\theta),\ldots,\sigma_p(\theta)), \theta \in \Theta$, as in \eqref{eq:parametricsubmodel}. We formulate a general procedure for estimation of an unknown ground truth parameter $\theta_0\in\Theta$ given a sample $I_1(\omega_1),\ldots,I_1(\omega_N)$ of $N$ endpoint images sampled from the SEPDA equations with noise fields $\sigma_1(\theta_0),\ldots, \sigma_p(\theta_0)$.

To estimate the endpoint moment $\mom{I_1}(\theta)$ for a given choice of noise fields, let $\widetilde{\mom{I_1}}(\theta)$ denote the endpoint moment image obtained from solving \eqref{eq:momentequation2} with noise fields $\sigma_1(\theta),\ldots,\sigma_p(\theta)$. Then for any similarity measure $d\colon \Imspace\times \Imspace \to [0,\infty)$ of images, we can construct a corresponding loss function $\ell\colon \Theta \to [0,\infty)$ given by
\begin{equation}\label{eq:lossfunction}
    \ell(\theta) = d\Big(\widetilde{\mom{I_1}}(\theta), \frac{1}{N} \sum_{i=1}^N I_1(\omega_i)\Big).
\end{equation}
As $\ell(\theta)$ compares how similar the approximated moment $\widetilde{\mom{I_1}}(\theta)$ is to the empirical mean of the observed images $\frac{1}{N} \sum_{i=1}^N I_1(\omega_i)$, the difference should be small when $\theta=\theta_0$ since they are both approximations of $\mom{I_1}(\theta_0)$. Thus we are led to define the corresponding moment estimator
\begin{equation}\label{eq:thetahat}
    \hat{\theta} \coloneqq \argmin_{\theta \in \Theta} \ell(\theta).
\end{equation}
This estimator can be computed in practice as an optimization problem. 

One reasonable choice for similarity measure is the squared distance $d(I,I') = SSD(I,I')= \|I-I'\|_{L^2}^2$, which is often used for classical image registration. Another reasonable choice comes from the normalized cross-correlation (NCC) given by:
\begin{equation}
    NCC(I,I') = 
    \frac{\inp{I}{I'}_{L^2}}
    {\|I\|_{L^2} \|I'\|_{L^2}}, \qquad I,I'\in \Imspace,
\end{equation}
which has the corresponding similarity measure given by $d_{NCC} \coloneqq 1-NCC^2$. A desirable property of this similarity measure is that it is invariant to changes in global pixel intensity. 

\subsection{Template estimation}
The set of parameters $\Theta$ can be expanded to include information beyond the noise fields. A natural addition would be to include the initial image $I_0$ in the parameter set and estimating it by minimizing the loss \eqref{eq:lossfunction} where gradients will now include image information. This will be analogous to template estimation as often employed together with image registration algorithms \cite{joshiUnbiasedDiffeomorphicAtlas2004}. Other possibilities include estimation of parameters of the LDDMM metric. Here, we focus on the noise structure.

\section{Parametric Submodels}\label{sub:models}
In this section, we discuss possible choices of parametric submodels. These particular parametrizations will be used in the simulations studies introduced in the next section. 

The choice of parametric submodel is a question of modeling and not a strict mathematical issue. To model the noise in a flexible manner, we first consider lattices of radially symmetric noise fields as in \cite{arnaudon2019geometric,kuhnel2018stochastic}. To be more precise, let $k\colon [0,\infty) \to [0,\infty)$ be a decreasing function and let $\Lambda = \{\mu_1,\ldots,\mu_p\}\subset (0,1)^2$ be a lattice of $p$ points. Then the corresponding parametrization of the noise fields is of the form
\begin{equation}\label{eq:GeneralFields}
    \sigma_{lm} (x,y) 
    = \lambda_l k(\|(x,y)-\mu_{l}\| / \tau)\mathbf{e}_m,
\end{equation}
for $m=1,2,$ and $l=1,\ldots,p$, and 
where $\lambda_1,\ldots,\lambda_p$ are amplitude parameters, $\tau$ is the width of the noise fields and $\mathbf{e}_1$ and $\mathbf{e}_2$ are the standard basis vectors in $\real^2$. Thus at each lattice point $\mu_l$, there are two corresponding noise fields $\sigma_{l1}$ and $\sigma_{l2}$ which are going to contribute with independent driving noises $W_t^{l1}$ and $W_t^{l2}$. 
A first choice for $k$ could be the Gaussian kernel $k(x) = (2\pi)^{-1/2}e^{-x^2/2}$, as considered in \cite{arnaudon2019geometric}, and we refer to such noise fields as \textit{Gaussian noise fields}. However, to avoid interference between noise fields, another reasonable choice is the cubic B-spline $k=\beta_3$,
\begin{equation*}
    6\beta_3(x) 
    = (x+2)_{+}^3 - 4(x+1)_+^3 + 6(x)_+^3-4(x-1)_+^3+(x-2)_+^3,
\end{equation*}
which has the advantage of having compact support. Here $(\cdot)_+ \coloneqq \max(\cdot, 0)$ denotes the positive part.

The simulations, introduced in the next section, were initially run with a $3\times 3$ square lattice as in \cite{arnaudon2019geometric,kuhnel2018stochastic}, but later extended to a $4\times 4$ lattice to get a more uniform cover of the domain. The explicit parametrization of the square lattice of Gaussian noise fields used in the simulations is:
\begin{equation}\label{eq:SquareLatticeGauss}
    \sigma_{ijm}(x,y) 
    = \lambda_{ij}\exp\left (-\frac{1}{2\tau^2}\|(x,y)-\mu_{ij}\|^2\right ) \mathbf{e}_m 
\end{equation}
for $m=1,2,$ and $1\leq i,j\leq 4$ and with $\mu_{ij} = (i/5,j/5)$. 

\begin{figure}[htb!]
    \centering
    \includegraphics[width = \linewidth]{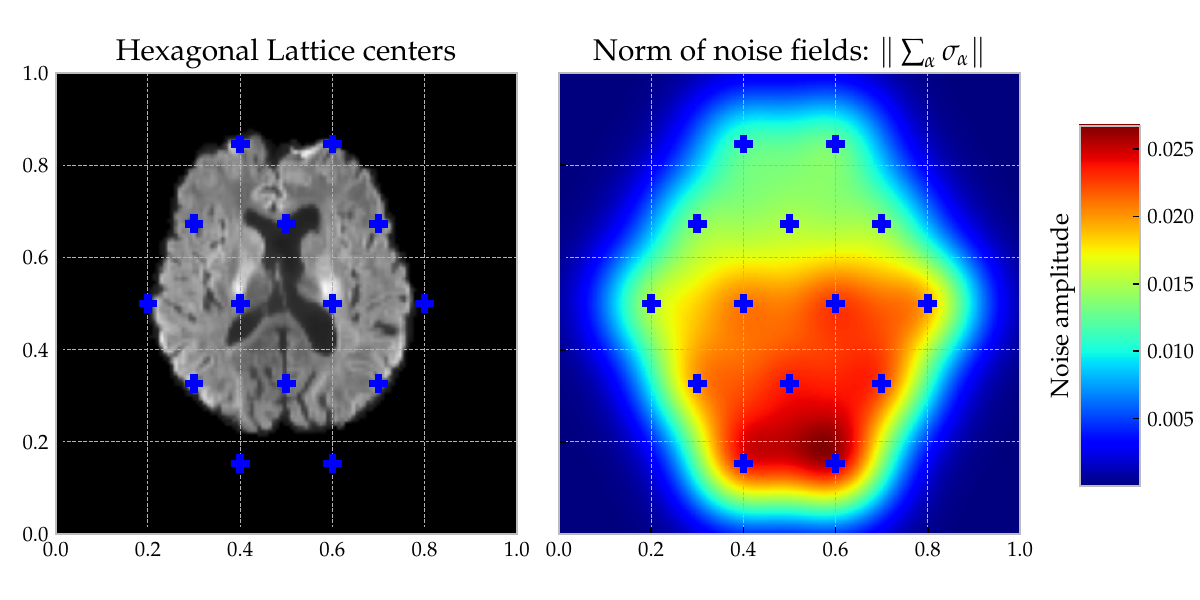}
    \caption{The left plot shows the hexagonal lattice placed on top of the brain MR-slice. The right plot shows the norm of a particular choice of Gaussian noise fields centered around the hexagonal lattice points.}
    \label{fig:HexLattice}
\end{figure}

To obtain a more dense cover, we may also consider a \emph{hexagonal} (triangular) lattice. The hexagonal lattice structure leads to the densest packing of congruent circles in the plane, which is suitable for the radially symmetric noise fields. Moreover, the simulations were performed on a brain MR-slice concentrated at the center of $(0,1)^2$, which the square lattice does exploit. In this view, a symmetrically centered hexagonal lattice of 14 gridpoints, shown in Figure~\ref{fig:HexLattice}, was used.  

Another parametrization considered was the \emph{sinusoidal noise fields} given by
\begin{equation}\label{eq:sinusoidal}
    \sigma_{nm\ell}(x,y) = c_{nm}\sin(n\pi x)\sin(m\pi y)\mathbf{e}_\ell,
\end{equation}
for $n,m = 1,\ldots, q$, and $\ell=1,2.$ The motivation for such noise fields comes from the fact that a sufficiently regular function $f\colon [0,1]^2 \to \real$ with boundary condition $f|_{\partial [0,1]^2}=0$ can be written as a sine series:
\begin{equation}\label{eq:sinedecomp}
    f(x,y) = \sum_{n,m=1}^\infty c_{nm} \sin(n\pi x)\sin(m\pi y).
\end{equation}
\section{Simulation Studies}
\label{sec:results}
In this section, we present the results of the following two simulation studies.

\paragraph{Experiment A}
Here the objective is to test the identifiability of $\theta$ for various parametric submodels $\Sigma(\theta)$ using the moment estimator $\hat \theta$.
The general experimental procedure is therefore:
\begin{enumerate}
    \item Consider a parametric model $\Sigma(\theta), \theta \in \Theta$, of the noise fields and a ground truth parameter $\theta_0\in \Theta$.
    \item Sample $N$ i.i.d. endpoint images $\{I_1(\omega_i)\}_{i=1,\ldots,N}$ by numerically integrating the SEPDA equations with noise fields $\Sigma(\theta_0)$.
    \item Estimate $\theta_0$ with the estimator $\hat \theta$ by optimizing over $\ell(\theta)$ for the synthetic dataset, and compare $\hat \theta$ with $\theta$.
\end{enumerate}

\paragraph{Experiment B}
Here the objective is to compare the estimated noise fields $\sigma_\alpha(\hat \theta)$ with the ground truth noise fields for a misspecified parametrization $\Sigma(\theta)$. Thus the procedure is the following:
\begin{enumerate}
    \item Consider ground truth noise fields $\varsigma_1, \ldots, \varsigma_\varrho \in \VF$ and a parametric model of noise fields $\Sigma(\theta),\theta \in \Theta$, such that $(\varsigma_1,\ldots,\varsigma_\varrho) \notin \Sigma(\Theta)$.
    \item Sample $N$ i.i.d. endpoint images $\{I_1(\omega_i)\}_{i=1,\ldots,N}$ from the SEPDA equations with $\varsigma_1, \ldots, \varsigma_\varrho$ as noise fields.
    \item Compute the estimator $\hat{\theta}$ given in \eqref{eq:thetahat} from the synthetically sampled dataset and compare $\Sigma(\hat \theta)$ with $\varsigma_1, \ldots, \varsigma_\varrho$.
\end{enumerate}

\subsection{Implementation Details}
\label{sec:implementation}

To conduct the simulation studies, an implementation\footnote{\url{https://github.com/AlexanderChristgau/mermaid}} was written based on the preexisting \mermaid library\footnote{\url{https://github.com/uncbiag/mermaid}}. The \mermaid toolbox contains various image registration methods via automatic differentiation in \pytorch \cite{paszke2017automatic}, including a shooting method based on the EPDiff equation \eqref{eq:EPDiff} and the image advection equation \eqref{eq:adveceq}. 

To simulate data from the SEPDA equations, a Heun scheme \cite{platen2010numerical} for Stratonovich integration was implemented into the \mermaid script.

To estimate the noise field parameter, the approximated moment equations - \eqref{eq:momentequation1} and \eqref{eq:momentequation2} - were implemented similarly to the existing registration models. In particular, the implementation uses the \torchdiffeq library for integration with the \emph{adjoint method} \cite{chen2018neuralode}. This enables the gradient of the loss to be computed without memory issues, and thus the built-in \pytorch optimizers can be applied to approximate $\hat{\theta}$ from \eqref{eq:thetahat}. For consistency, Adaptive Moment estimation (ADAM) was used throughout all simulation studies.

The simulations were performed on the brain MR-slices shown in Figure \ref{fig:image_registration}. In this case the image domain was discretized as $\widetilde{\mathcal{D}} = \{(i,j)/127\}_{0\leq i,j\leq 127}$, but the implementation is not specific to $\widetilde{\mathcal{D}}$. The simulations were conducted with an i5 processor (1.80GHz × 4) and thus, with the availability of CUDA in the \mermaid library, it should be possible to upscale to 3D images in future research.

\subsection{Experiment A}
For the lattice parametrizations, the ground truth amplitudes were chosen such that the resulting noise was on a realistic scale of the brain image, and such that the parameters were spread out, allowing for improved visualization. The ground truth for the Gaussian noise fields was given by lexicographically ordering the lattice points and setting $\lambda_i^{\text{Gauss}} = 0.005 + 0.000625\cdot (i+2\sin i)$. For the cubic B-splines the ground truth was given similarly by $\lambda_i^{\text{B-spline}} =\lambda_i^{\text{Gauss}}/5$. The kernel width $\tau$ was specified as a hyperparameter such that the noise fields had an appropriate overlap. For example, Figure \ref{fig:HexLattice} shows the resulting scalar \emph{norm field} $\|\sum_\alpha \sigma_\alpha\|$ of the ground truth Gaussian hexagonal lattice with $\tau^2 = 0.008$. Otherwise the width was specified as $\tau=0.15$ and $\tau = 0.1$ for the B-spline noise fields and the Gaussian square lattice, respectively. 

For each parametric submodel a dataset of $500$ images was sampled by numerically integrating the SEPDA equations with the Heun scheme using 128 equidistant time steps. 

\begin{figure*}
    \centering
    \includegraphics[width = .9\linewidth]{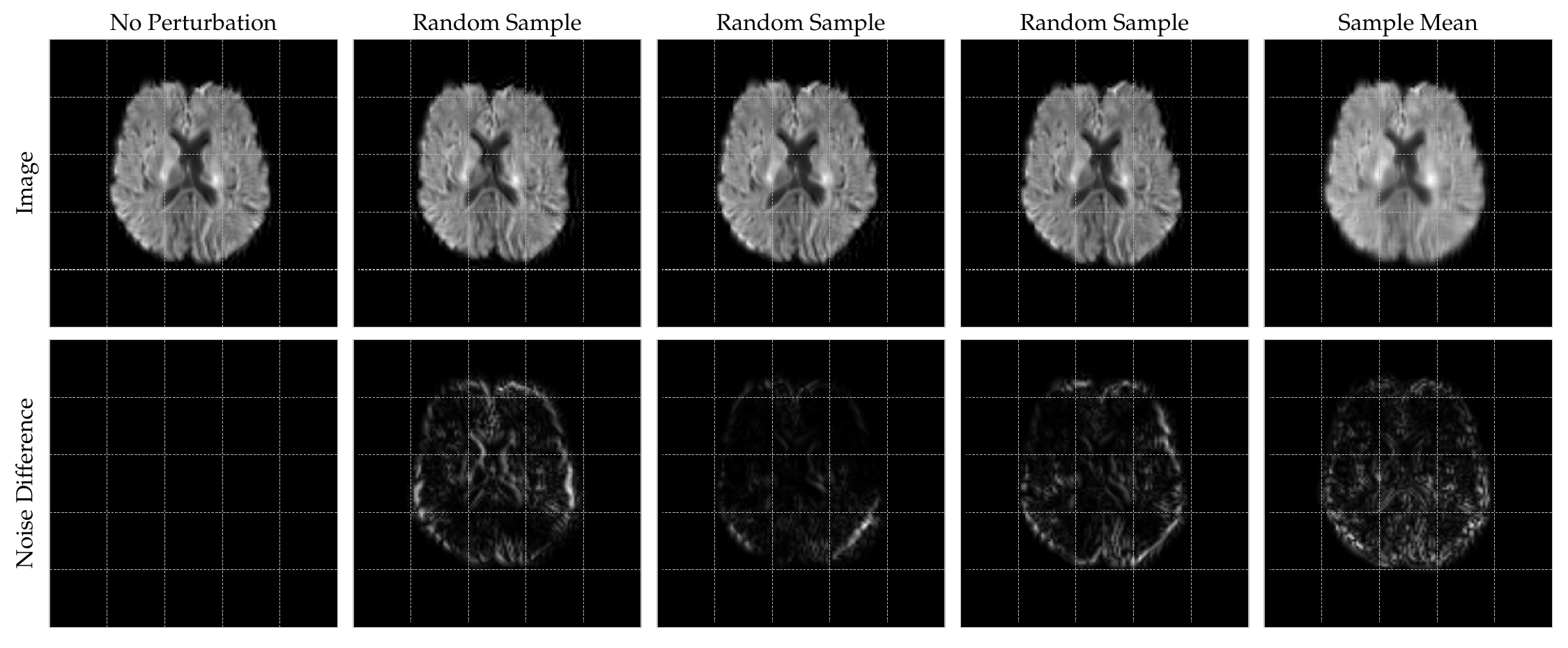}
    \caption{
    The rows show the endpoint images and their difference from the deterministic endpoint, respectively, for three random samples and the sample mean of a synthetic dataset sampled from the SEPDA equations. Here the noise fields are Gaussian and arranged in a hexagonal lattice.}
    \label{fig:HexGaussSamples}
\end{figure*}

Figure \ref{fig:HexGaussSamples} shows three random samples and the sample average $\hat I_1 \coloneqq \frac{1}{500} \sum_{i=1}^{500} I_1(\omega_i)$ for a dataset obtained from the Gaussian hexagonal lattice shown in Figure \ref{fig:HexLattice}. Note that while the variation is subtle, it is greater towards the bottom of the brain in accordance with the specification of the noise fields.

Once each dataset was sampled, the estimator $\hat \theta$ from \eqref{eq:thetahat} could be obtained by minimizing the loss function  $\ell(\theta) = \ell(\lambda_{11},\ldots,\lambda_{44}) = d(\widetilde{\mom{I_1}}, \hat I_1)$. During initial experimentation the NCC-similarity was found to be more effective for parameter estimation, and hence $d=d_{NCC}$ was chosen for all the simulations. 

\begin{figure}[htb!]
    \centering
    \includegraphics[width = \linewidth]{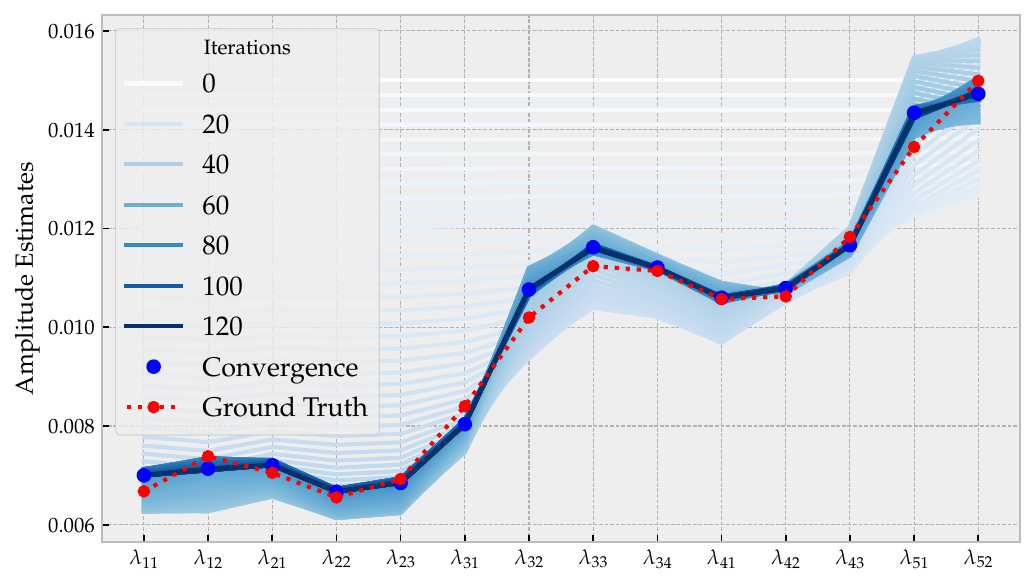}
    \caption{Parameter inference performed on a dataset generated by the Gaussian hexagonal lattice showing convergence towards an accurate estimate of the ground truth (red).}
    \label{fig:HexGaussInference}
\end{figure}

Figure~\ref{fig:HexGaussInference} and Figure~\ref{fig:HexBsplineInf} illustrate the optimization procedures for the Gaussian noise fields and the B-spline noise fields, respectively, both arranged in the hexagonal lattice. The plots are similar, and in both cases the optimization procedure converges towards an accurate estimate of the ground truth after approximately 80 iterations. 

The results for the square lattice, shown in Figure~\ref{fig:4x4GaussInferenceWithoutT}, were decent but inferior to those of the hexagonal lattice. Observe that estimates of corner amplitude parameters such as $\lambda_{44}$ are more inaccurate. A reasonable explanation is that noise fields supported outside of the brain are less constrained by the loss function, and thus they may depart from the ground-truth more significantly. 

Initially the inclusion of $\tau^2$ as a parameter was also tried and the corresponding results are shown in Figure~\ref{fig:4x4GaussInferenceWithTau}. The inclusion did not seem to affect the relative magnitudes of the amplitude parameters in comparison to Figure~\ref{fig:4x4GaussInferenceWithoutT}. However, a slight decrease across all amplitude parameters. Intuitively, this makes sense as the width, $\tau^2$, was over estimated and since the width and the amplitude control the magnitude of the noise together. In this view, $\tau^2$ was kept fixed as a hyperparameter for the other experiments.

For the sinusoidal noise fields the ground truth was selected by first considering the function 
\begin{equation*}
    f\colon (x,y) \mapsto xy^2(1-x)(1-y) \cos(5x)\cos(5y).
\end{equation*}
Then the ground truth was chosen from the sine basis coefficients of $f$ given by
\begin{equation*}
    c_{nm} 
    = 4 \int_0^1 \int_0^1 f(x,y)\sin(n\pi x)\sin(m\pi y)dxdy,
\end{equation*}
for $1\leq n,m \leq 4$. The same procedure as in the preceding experiments was followed and the results for the $f$-frequencies are found in Figure \ref{fig:FourierInferenceOfFunc}.
Note that while the general scale of the estimates seems reasonable, the accuracy of each estimate is poor when compared to the lattice parametrizations. As a consequence, the estimates of the frequency coefficients cannot be used to reconstruct an approximation of the function $f$.

\begin{figure*}[htb!]
    \centering
    \includegraphics[width = .9\linewidth]{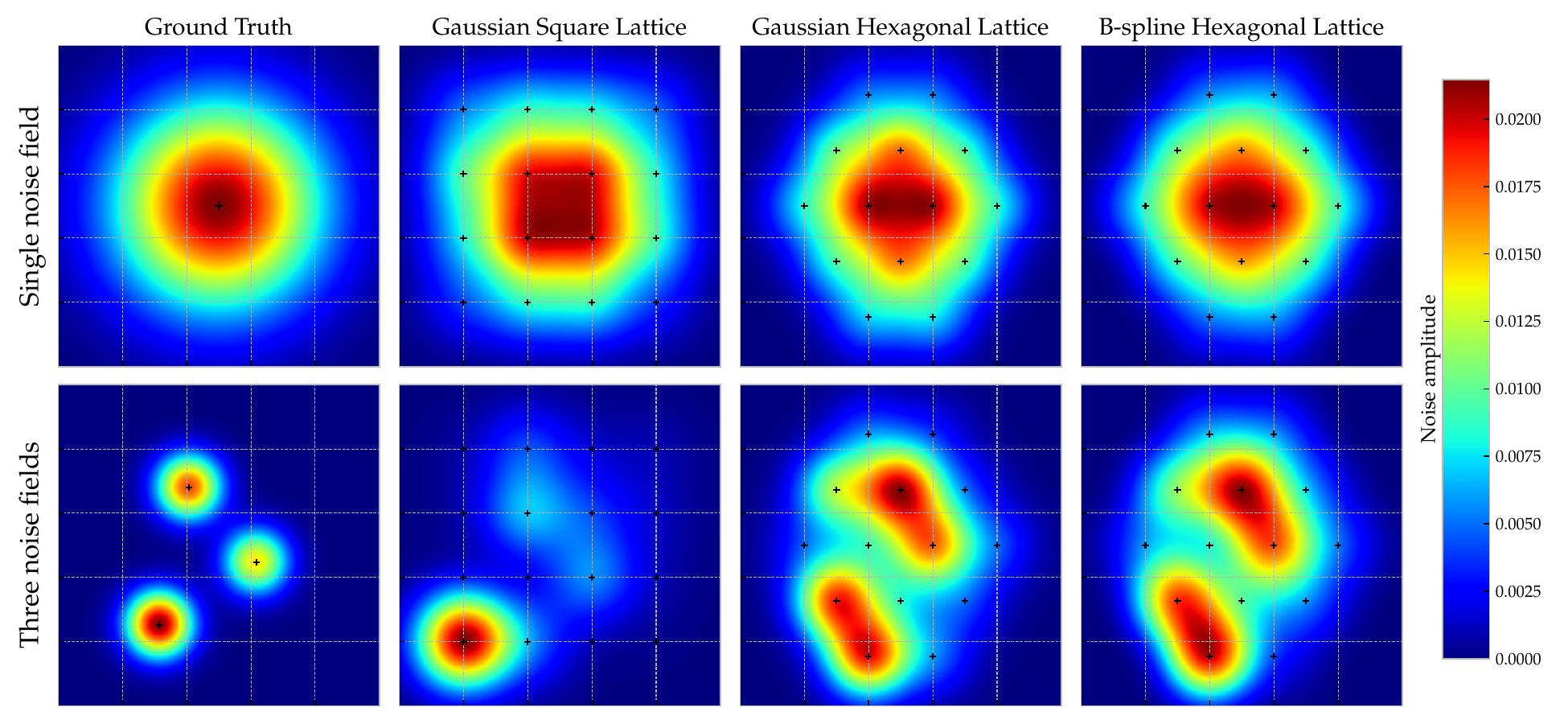}
    \caption{Noise field inference under model misspecification. The top row shows the models fits onto a single large noise field while the bottom row shows the model fits onto three smaller distinct noise fields.}
    \label{fig:ExperimentB}
\end{figure*}

To end experiment A, Table \ref{tab:RelativeErrors} shows the relative errors of the estimates given by $\|\theta_0-\hat\theta\|/\|\theta_0\|$.
\begin{table}[htb!]
    \centering
    \begin{tabular}{|l|r|}
        \hline
         Parametrization & Relative error \\ \hline
         Gaussian hexagonal & 0.0314 \\ \hline
         B-spline hexagonal &{\bf 0.0262} \\ \hline
         Gaussian square & 0.0795 \\ \hline
         Sinusoidal & 0.320 \\ \hline 
    \end{tabular}
    \caption{Relative error $\|\theta_0 -\hat\theta\|/\|\theta_0\|$ for various parametric sub-models.}
    \label{tab:RelativeErrors}
\end{table}

We note that the hexagonal lattice with cubic B-splines provides the most precise estimates, followed by the hexagonal Gaussian noise fields. This is presumably because the compact support of the B-spline leads to less interference between the noise fields.  Interference between noise fields may also explain the sinusoidal noise fields poor performance. Inclusion of higher order moments in future research could improve separation of interfering noise fields. 
 We note also that the hexagonal lattice solves the issue of corner noise fields having insufficient overlap, as it was designed for these particular images with a brain placed at the center. For other images containing a different shape, it may be more appropriate with another configuration of noise fields.

\subsection{Experiment B}

The ground truth noise fields used for sampling the synthetic datasets were
\begin{enumerate}
    \item A single Gaussian noise field placed at the center.
    \item Three randomly placed Gaussian noise fields\footnote{Their coordinates are (0.617, 0.447), (0.406, 0.681) and (0.314, 0.251).}.
\end{enumerate}
One approach to comparing the fitted noise fields $\Sigma(\hat \theta)$ with the ground truth noise fields is to display their respective norm fields $\|\sum_\alpha \sigma_\alpha\|$ perform a visual inspection.

Figure \ref{fig:ExperimentB} shows the single Gaussian noise field in the top-left subplot and the three distinct Gaussian noise fields in the lower-left subplot. The remaining subplots show the norm fields of the respective models fitted onto a corresponding synthetic dataset of $500$ images. 
We observe that the fitted models visually resemble the ground truth noise fields despite being restricted to their lattice structure with fixed width. For the single noise field the resemblance is clear for all the fitted models and it is ambiguous if any model is better. For the three smaller noise fields the resemblance is slightly less clear. The square lattice overestimates the noise field placed in the lower left and underestimates the two other noise fields, while both of the hexagonal lattice models are unable to separate the noise fields towards the upper right (another visualization of this is given in Figure \ref{fig:Meandiff} in the appendix).
To get a numerical comparison, consider the quantity
\begin{equation}\label{eq:noisefield_error}
    SSD\left (\Big \|\sum_\alpha \sigma_\alpha(\hat \theta) \Big \|, \Big \|\sum_\alpha \varsigma_\alpha\Big\|\right),
\end{equation}
i.e. the squared deviation of the fitted noise fields from the ground truth. Table \ref{tab:noisefielderrors} shows the above quantity for the various fitted models. 

\begin{table}[htb!]
    \centering
    \begin{tabular}{c||c|c|c}
        Ground truth & Gauss. sq. & Gauss. hex. & B-spline  \\ \hline
        Single Gaussian & 0.227 & 0.128 & {\bf 0.124} \\ \hline
        Three Gaussian & 1.21 & {\bf 0.454} & 0.751
    \end{tabular}
    \caption{The similarity between the ground truth and fitted noise fields measured by SSD-similarity \eqref{eq:noisefield_error}.}
    \label{tab:noisefielderrors}
\end{table}
We observe that indeed the single Gaussian noise field is easier to approximate than the three smaller ones. The square lattice yields the worst approximations, which is consistent with the visual inspection and the findings of Experiment A.

In summary, the example in the top row of Figure~\ref{fig:ExperimentB} is well resolved with our choice of noise fields, whereas the bottom row is not.
These issues may be dealt with by increasing the density of noise fields, i.e., making the lattices finer or by optimizing the centers of the noise fields in addition to their amplitudes.
In an ideal situation, such parametrized noise field positions should adapt to the shape of the noise, but, in practice, they may require repulsion forces between them, with a risk of overfitting. 

\section{Conclusions and Future Directions}

Based on the stochastic generalization of the LDDMM framework introduced in~\cite{arnaudon2019geometric}, and in particular its specialization to images, we derived the first order moment approximations of the SEPDA equations to construct an estimator for the noise field parameters. 
With the proposed model for estimation of time-continuous variation in deformations of shapes captured in images, we carried out two main numerical experiments on medical imagery.\\
In the first experiment, we found that noise field estimators are able to accurately predict the ground truth, provided that the noise fields have sufficient overlap with the content of the image but not too much between themselves, indicating when this method works and how to specify noise field locations.

In the second experiment, we found that the parametric submodels are able to approximate some unknown ground truth noise fields. However, the accuracy of the approximations were, perhaps unsurprisingly, limited to the resolution of the grids. Therefore, it could be interesting to experiment with more refined grids in future work, as well as different kinds of generating noise fields. 

Beyond the simulation studies, a natural next step is applying our method to estimate random variation located in the deformation of shapes such as human organs. For such real datasets, the questions of how many noise fields and where to place them on the image may become important for accurate noise estimations while remaining computationally tractable. 
In particular, one could attempt to optimize for the position and width of all or some of the noise fields to make this method more flexible and with less parameters to set from the users.

Other possible directions for future research include the possibility of time-dependent noise fields for specific applications, other types of profile of noise fields or the use of higher order moment approximations of the SEPDA equations, with e.g. the cluster-expansion approach, to tackle more challenging real world examples, such as longitudinal data extension to 3D images or diffusion tensor imaging. 
\begin{acknowledgements}
AMC is grateful to Zhengyang Shen for helpful discussions and guidance for using the \mermaid library. The work presented is supported by the Villum Foundation grant 00022924, and the Novo Nordisk Foundation grant NNF18OC0052000. 
\end{acknowledgements}

%
%
\bibliographystyle{spmpsci}      
\bibliography{bibliography.bib}

\clearpage
\appendix
\begin{figure}[!b]
\section{Additional Figures and Plots}
    \onecolumn
    \begin{minipage}{\textwidth}
        \centering
        \includegraphics[width=0.9\textwidth]{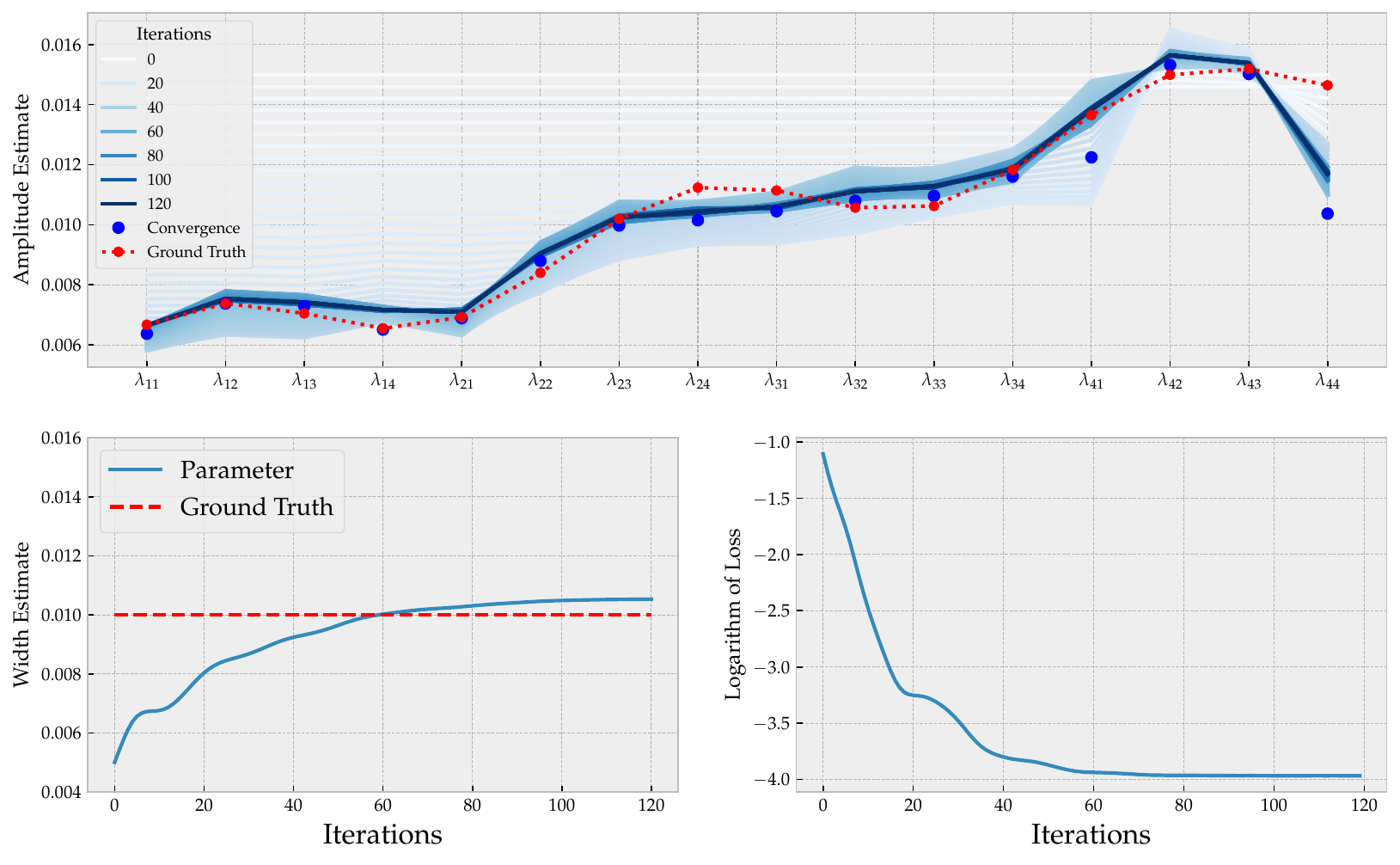} 
        \caption{
        Optimization procedure of $\ell(\theta) = d(\widetilde{\mom{I_1}}, \hat I_1)$ used to infer amplitude parameters $\lambda_{ij}$ and width parameter $\tau^2$ for a $4\times 4$ lattice of Gaussian noise fields given by \eqref{eq:SquareLatticeGauss}.}
        \label{fig:4x4GaussInferenceWithTau}
    \end{minipage} %
    
    \begin{minipage}{.47\textwidth}
        \centering
        \includegraphics[width = \textwidth]{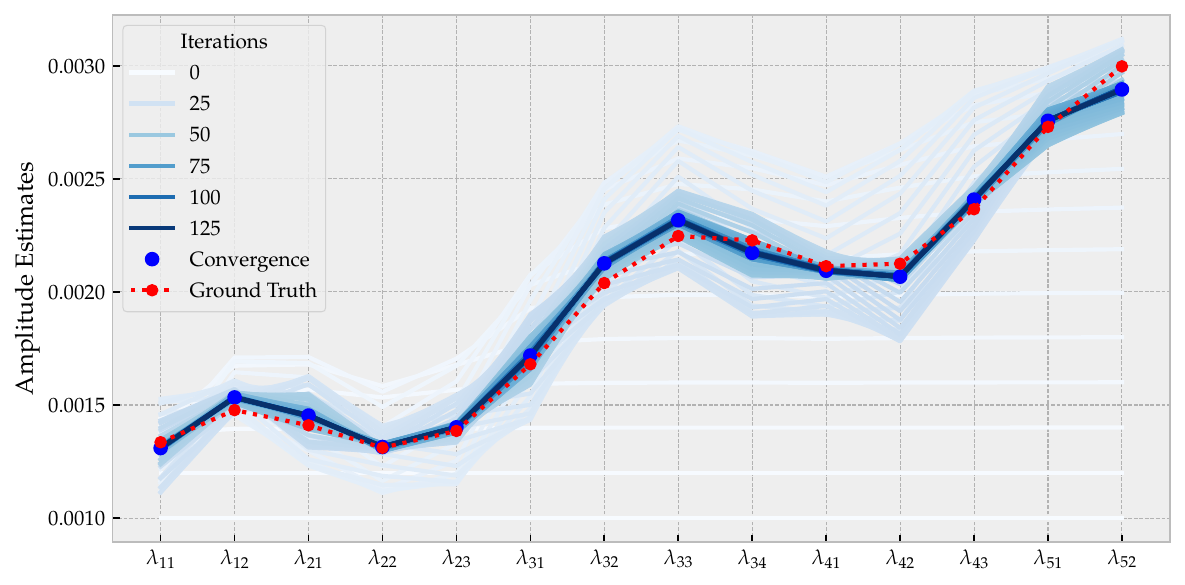}
        \caption{Parameter inference for data generated by B-splines noise fields.}
        \label{fig:HexBsplineInf}
    \end{minipage} 
    \begin{minipage}{.47\textwidth}
        \centering
        \includegraphics[width = \textwidth]{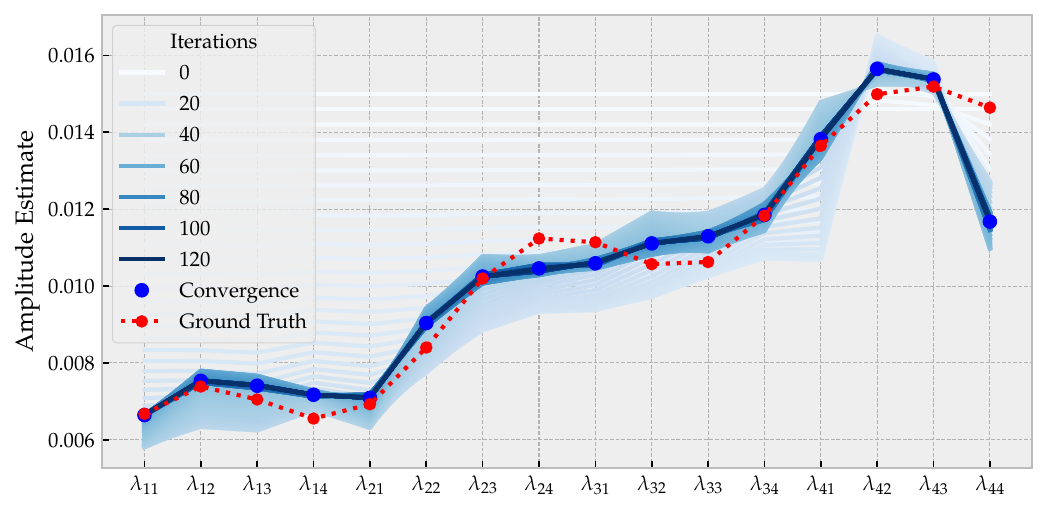}
        \caption{
        Optimization procedure of $\ell(\theta) = d(\widetilde{\mom{I_1}}, \hat I_1)$ used to infer amplitude parameters $\lambda_{ij}$ for a $4\times 4$ lattice of Gaussian noise fields, with the width ($\tau^2 = 0.01$) considered as fixed.}
        \label{fig:4x4GaussInferenceWithoutT}
    \end{minipage} 
    \begin{minipage}{.47\textwidth}
        \centering
        \includegraphics[width = \textwidth]{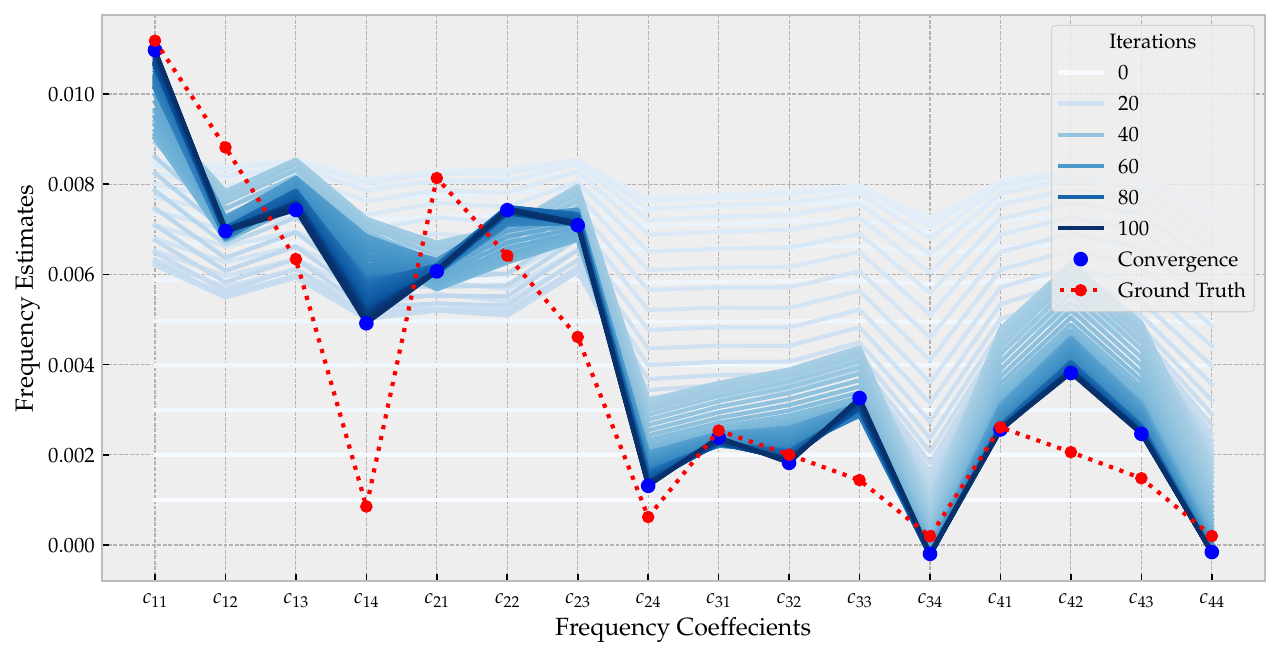}
        \caption{Inference of frequency coefficients of sinusoidal noise fields.}
        \label{fig:FourierInferenceOfFunc}
    \end{minipage}
\end{figure}

\begin{figure*}[htpb]
    \centering
    \includegraphics[width = 0.9\textwidth]{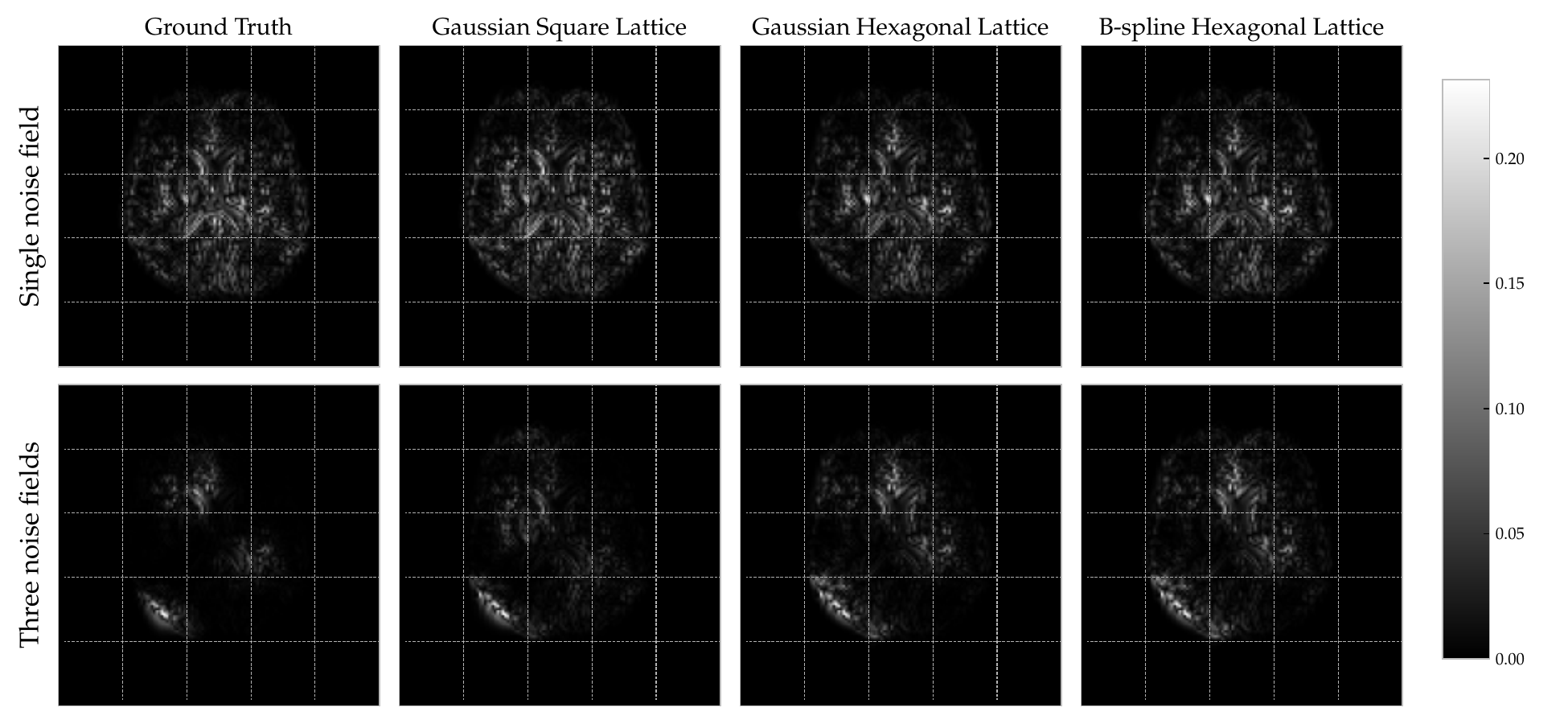}
    \caption{The first column shows the mean noise difference $|\hat I_1 -I_1^{\text{deterministic}}|$ for the the datasets sampled in Experiment B. The remaining columns show the corresponding predicted mean noise differences $|\widetilde{\mom{I_1}}(\hat \theta)-I_1^{\text{deterministic}}|$ for the various model fits.}
    \label{fig:Meandiff}
\end{figure*}

\end{document}